   \documentclass[letterpaper, 10pt, twocolumn]{article}
   \usepackage{clean}
   \shorttitle{Teaching contact-rich from vision-only demonstrations}
 \newcommand{\CLASSINPUTtoptextmargin}{19.1mm}
 \newcommand{\CLASSINPUTbottomtextmargin}{19.1mm}
 \newcommand{\CLASSINPUTinnersidemargin}{19.1mm}
 \newcommand{\CLASSINPUToutersidemargin}{19.1mm}
 
\usepackage{graphicx}
\usepackage{subfig}
\usepackage{amsmath,amssymb,amsfonts}
\usepackage{algorithm2e}
\usepackage{xcolor}
\usepackage[colorlinks=true, linkcolor=black, urlcolor=cyan, filecolor=black, citecolor=black]{hyperref}
\usepackage{url}
\usepackage{amsmath,bm}
\usepackage{multicol, multirow, makecell} % For table magic
\usepackage{comment}

\usepackage{cite}

\title{Teaching contact-rich tasks from visual demonstrations \\[0.15cm] by constraint extraction}
\author{ Christian Hegeler$^1$, Filippo Rozzi$^2$,  Loris Roveda$^3$, and Kevin Haninger$^1$ %\vspace{-30pt} % @Kevin 30.03.23 Was needed to make the IEEE margins
\thanks{$^1$Department of Automation at Fraunhofer IPK, Berlin, Germany}
\thanks{$^2$Politecnico di Milano, Department of Mechanical Engineering, Milano, Italy}
\thanks{$^3$Istituto Dalle Molle di Studi sull'Intelligenza Artificiale (IDSIA), Scuola Universitaria Professionale della Svizzera Italiana (SUPSI), Università della Svizzera Italiana (USI) IDSIA-SUPSI, Lugano, Switzerland {\tt loris.roveda@idsia.ch}}
\thanks{Corresponding author: {\tt Christian-Hegeler@proton.me}. This project has received funding from the European Union's Horizon 2020 research and innovation programme under grant agreement 101058521 — CONVERGING.}}

\begin{document}

\maketitle
\begin{abstract}
Contact-rich manipulation involves kinematic constraints on the task motion, typically with discrete transitions between these constraints during the task. Allowing the robot to detect and reason about these contact constraints can support robust and dynamic manipulation, but how can these contact models be efficiently learned? Purely visual observations are an attractive data source, allowing passive task demonstrations with unmodified objects. Existing approaches for vision-only learning from demonstration are effective in pick-and-place applications and planar tasks. Nevertheless, accuracy/occlusions and unobserved task dynamics can limit their robustness in contact-rich manipulation. To use visual demonstrations for contact-rich robotic tasks, we consider the demonstration of pose trajectories with transitions between holonomic kinematic constraints, first clustering the trajectories into discrete contact modes, then fitting kinematic constraints per each mode. The fit constraints are then used to (i) detect contact online with force/torque measurements and (ii) plan the robot policy with respect to the active constraint. We demonstrate the approach with real experiments, on cabling and rake tasks, showing the approach gives robust manipulation through contact transitions.
\end{abstract}

\section{Introduction}\label{Introduction}

Contact modeling and control are essential for robots to physically interact with their environments \cite{suomalainen2022survey}. When the environment is less structured, the contact conditions will vary and the ability to detect and reason about contacts is needed for robust and dynamic manipulation, such as for drawers, doors, and cables \cite{bicchi2000robotic}. 

Contact constraints are important for the safety, performance, and robustness of manipulation \cite{smith2012dual,jimenez2012survey,li2019survey}. For safety, violating the current constraint can result in excessive force and transitions between contact with excessive velocity can result in collision forces. For performance, constraints can be exploited to reduce variation in trajectories, and trajectory planning with respect to the constraints can also exploit dynamics (\textit{e.g.}, Coulomb friction, constraint-space inertia). For robustness, changes in contact modes can be used to monitor a task to identify task success, failure, or other discrete modes. Methods exist to plan through constrained motion, but they need models of the constraints.

\subsection{Related works}

Contact models - \textit{e.g.}, the contact normal or signed distance function - are key for modern planning-based approaches to locomotion and contact-rich manipulation \cite{pang2022}. These are application specific and often extracted from CAD models. However, they can also be learned from demonstrations, such as by using an instrumented tool that measures the pose and forces of a human \cite{subramani2018}. 

\begin{figure}
    \centering
    \includegraphics[width=0.43\textwidth]{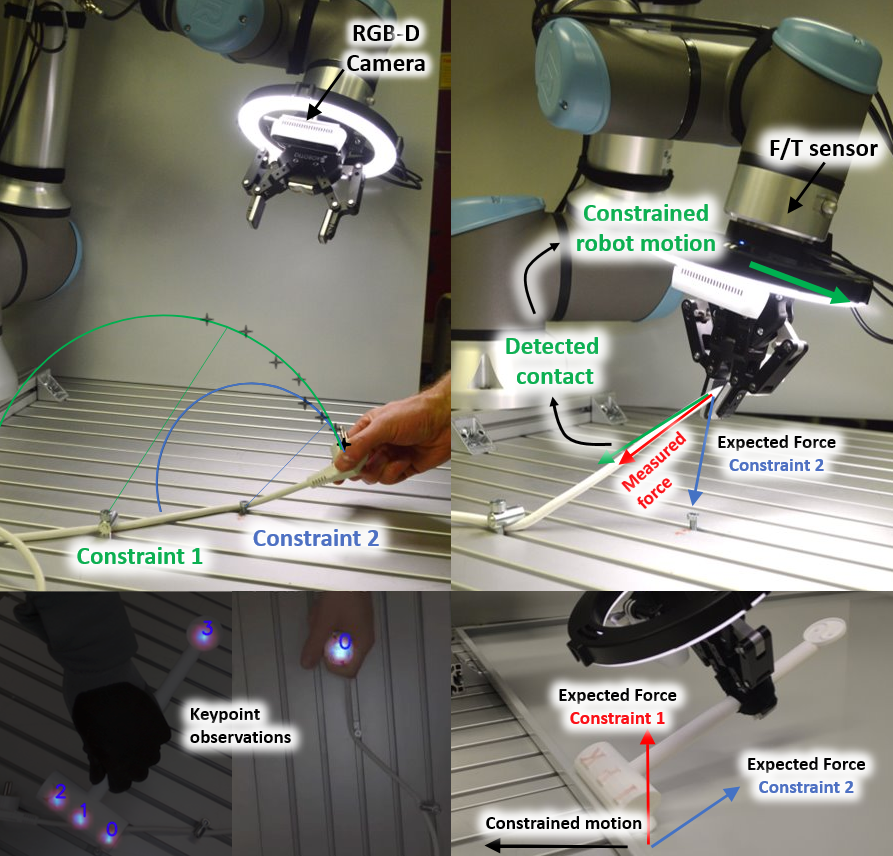}
    \caption{Setup of the approach, where the keypoint trajectory of an object is tracked by an RGBD camera (left), the data are clustered, and kinematic constraints are fit. The constraint model is then exploited by the robot online (right), where measured force is compared with the constraint Jacobians to identify constraint online and the robot's trajectory is adapted to enter and maintain constraints.}
    \label{fig:overview}
\end{figure}

Learning from demonstration (LfD) is a powerful tool, developing methods to specify robot tasks in an efficient and natural way \cite{ravichandar2020}. Classical LfD approaches such as dynamic movement primitives can learn and adapt free-space trajectories to new start and goal positions, and have also been extended to consider contact \cite{denisa2016}. However, demonstrating force trajectories can be complex: either a 2nd force/torque sensor is needed \cite{tang2015, chang2022}, or the robot must be teleoperated \cite{vecerik2018, scherzinger2019}, limiting ease of demonstration and dexterity in fine tasks, respectively. Furthermore, this approach must balance whether the force or position trajectory should be tracked, a question of the robot's impedance \cite{stulp2012}. This can be optimized when the robot reproduces the trajectories \cite{chatzilygeroudis2019, chang2022}, however, introducing additional complexity in the definition of the reward function, data collection, and implementation. 

For LfD methods which are variance- \cite{franzese2021, haninger2022c} or stiffness-aware \cite{abu-dakka2018}, the task representation is continuous, not explicitly representing changes in the contact conditions. Other methods which learn discrete modalities from demonstration such as \cite{su2018, wang2019a} do not build explicit contact models, limiting the ability to use force information or planning methods.

Demonstrations can also be used with behavior cloning in deep learning-based approaches, initializing a task-specific policy \cite{florence2022}. This has been applied to contact-rich tasks like peg-in-hole insertion from visual information \cite{vecerik2018}. While some deep learning policies can incorporate force information \cite{lee2020c}, the ability of these methods to handle tighter tolerance tasks and extrapolate beyond the data has not yet been shown. In fact, the ability of standard neural networks to directly model contact is currently limited \cite{parmar2021}. 

\begin{figure}
    \centering
    \includegraphics[width=0.3\textwidth]{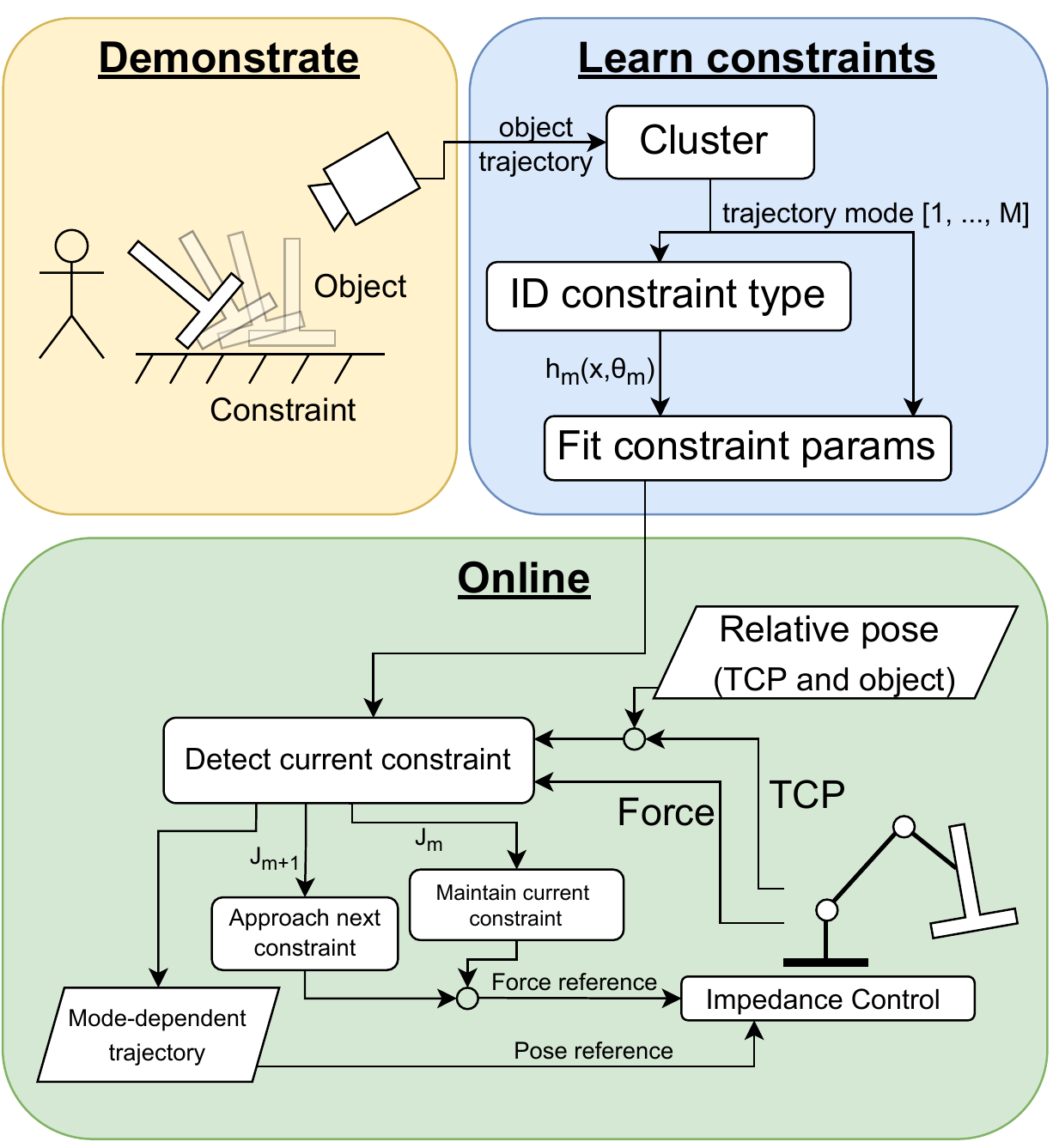}
    \caption{The components of the proposed approach, where the demosntrations and constraint learning are covered in Section III, and online control in Section IV.}
    \label{fig:process_description}
\end{figure}

\subsection{Paper Contribution}

In contact-rich manipulation tasks, object trajectories can be characterized by a set of key phases, depending on the type of constraint currently acting on the object. An autonomous robot should, therefore, be able to identify these constraints from such manipulation tasks. However, it is still difficult to get contact models for complex manipulation tasks, in particular avoiding the use of additional external sensors in the demonstrations. In addition, vision-only methods may lack robustness in reproducing contact-rich motion.

To address the above-mentioned issues, we consider using passive observations (\textit{i.e.}, developing a visual-only method) of human contact-rich demonstrations and note that, due to the noise and low stiffness of typical human motions, repeatability in the trajectories will be induced primarily by contact constraints. We show that, by segmenting the demonstrations and fitting the constraints, constraint-aware manipulation can be improved. This happens in two ways: i) using the kinematic constraints to derive the direction of expected force, allowing online classification of the contact condition from F/T measurements, and ii) using contact models for the planning process, setting a desired force to maintain the current constraint and approach the next one.

We validate this approach with real experiments, using keypoint detection on objects without markers, and reproducing contact-rich tasks. We find that clustering and fitting parameterized models is effective even on noisy RGBD data, and that online detection with constraint Jacobians is effective provided the Jacobians are not parallel at the current pose.

The main novelties of the approach proposed here approach are: (i) a validated method for extracting a discrete collection of geometric constraints from vision-only demonstrations; (ii) a method for the online detection of the active constraint, based on F/T measurements and the learned constraint geometry; (iii) a higher-level controller which sets the force reference of an impedance controller to exploit the constraint information to improve task robustness.

\subsection{Proposed Approach vs State of the Art}

Compared to related constraint-learning work \cite{subramani2018}, this method doesn't need additional motion markers or F/T sensors to collect demonstrations and the learned models are applied online. Compared with dense visual descriptors used in teaching by demonstration \cite{florence2022}, this method can handle contact-rich manipulation from human-readable visual information (pose from point correspondences). Compared with contact-DMPs (dynamic movement primitives) \cite{denisa2016}, the demonstrations are purely visual and produce explicit contact models. Compared with variance-aware \cite{franzese2021} or stiffness-aware \cite{abu-dakka2018} LfD, our method explicitly models constraints, and changes in contact, and can detect changes in contact online.

\section{Problem Statement and Approach}
This section formalizes the type of considered manipulation problem, where multiple kinematic constraints occur during its execution, then overviews the proposed approach. 

\subsection{Contact-rich Manipulation}
The contact-rich manipulation problem is expressed over the pose or local features of a manipulated object, $x\in\mathbb{R}^n$, where $n$ is the dimension of the position or pose observations. During a task, the object has a trajectory of $N$ points, $\mathcal{X} = [x_1, \dots, x_N]$. This trajectory has discrete changes in contact conditions, denoted sequentially as $m\in[1,\dots,M]$. Each mode has a holonomic constraint of the form $h_m(x)=0$, where $h:\mathbb{R}^n\rightarrow \mathbb{R}^{n_m}$ and $n_m$ is the degree of the constraint in $m$ \cite{acary2008}. A constraint $h_m(x)$ is enforced with constraint forces of the form $F=\lambda^TJ_m(x)$, where $F\in\mathbb{R}^n$ are the forces expressed in the coordinate system of $x$, $\lambda\in\mathbb{R}^{n_m}$ the Lagrange multipliers representing the magnitude of the force and $J_m = \partial h_m / \partial x$ the constraint Jacobian. 

\subsection{Problem Statement and Approach}
Given only pose trajectories $\mathcal{X}$, where a sequence of contact modes $[1, \dots, M]$ implicitly occur, it is desired that the robot can safely reproduce this sequence of contacts over small geometric variations in the constraints. 

The proposed approach can be seen in Figure \ref{fig:process_description}, where pose trajectories $\mathcal{X}$ are used to find kinematic constraint models. First, the data is clustered to produce $M$ datasets, corresponding to each constraint mode. Then, a constraint model is fit for each data cluster, where it is assumed that the constraint is one of a library of parameterized models $h_m(x, \theta_m)$, where the parameters $\theta_m$ are parameters fit to the clustered data. The fit models then give Jacobians $J_m$, which give the direction or subspace in which the mode's constraint forces are expected. Force measurements allow the identification of the constraint, and can inform the robot policy.  

\section{Visual Demonstrations}
This section describes how the visual demonstrations are collected and the data segmented. 
\subsection{Pose estimation}
\begin{figure}
    \centering
    \includegraphics[width=0.55\columnwidth]{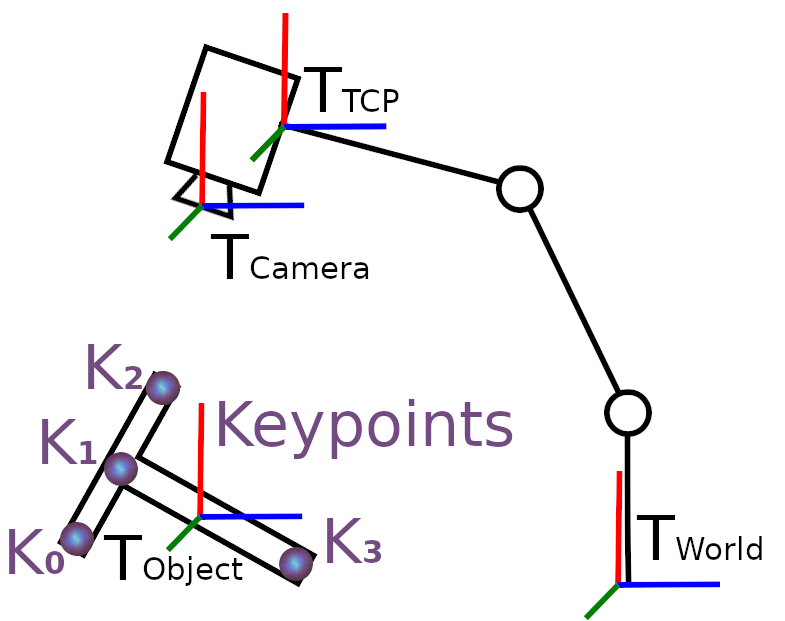}
    \caption{Overview of involved coordinate systems for object pose estimation}
    \label{fig:keypointframes}
\end{figure}

Poses of objects, such as the plug and rake seen in Figure \ref{fig:overview}, are estimated with the means of keypoints. Keypoints have a fixed relation to local visual object features and are ideally invariant to rigid transformations of object poses.
The keypoint location is estimated from RGB images in 2D pixel space based on the work of \cite{blomqvist2022} and are subsequently transformed into 3D world coordinates following the transformations shown in Fig. \ref{fig:keypointframes}. Using depth information and calibrated camera intrinsics, keypoints locations are lifted from pixel space into 3D space relative to the camera. The camera poses relative to world coordinates are estimated with the known tool center point (TCP) and the known fixed transformation from TCP to camera initially calibrated with the work of \cite{daniilidis1999hand}. 

Keypoints are detected using deep convolutional neural network models trained on object-specific data sets that consist of image data with corresponding keypoint labels. The image data is collected autonomously by the robot. The labeling process is semiautomatic, keypoints labeled in one image, then projected from pixel space to 3D world coordinates and back to pixel coordinates in the other images. When using three or more distinguishable keypoints, related object poses can be unambiguously described in $SE(3)$. For an initial keypoint configuration, an initial object pose is assigned with the center at the mean of all related keypoints and identity as rotation. When detecting keypoints for any new object pose, the pose is described by a transformation of the initial object pose that aligns both sets of keypoints. 

\subsection{Segmentation}\label{sec:segmentation}
%In contact-rich manipulation tasks, object trajectories can be characterized by a set of key phases, depending on the type of constraint currently acting on the object. An autonomous robot should therefore be able to identify these constraints from such manipulation tasks. For this, a segmentation approach should be considered as a core process to support the learning of contact constraints.
In this section we propose an autonomous segmentation framework to extract a set of segmentation points by dividing a multi-dimensional motion trajectory obtained from a single passive demonstration; the aim is to define a temporally ordered sequence of clusters in which just one constraint is present, such that each cluster can be used for the constraint learning. To this extent, we model the overall trajectory including the time component as a Gaussian Mixture Model (GMM) \cite{calinon2007} and we estimate the parameters of the model using the standard Expectation Maximization approach. Representing a multi-dimensional trajectory as a GMM provides a way of encoding the local direction and local relations (\textit{i.e.}, covariance) between the variables. The main advantage of such an estimation framework is that there is no need to manually predefine any internal parameters according to the types of given tasks and/or trajectories. 

We assume the ergodicity of the training data and we do not need to consider temporal and spatial variations between multiple demonstrations while preparing the training data. A multi-dimensional trajectory, $\mathcal{X}\in\mathbb{R}^{(D+1)\times N}$, is extracted from a passive observation. Here, $(D+1)$ denotes the $D$-dimensional spatial variables (\textit{e.g.}, object pose, velocity, keypoints coordinates) and the one-dimensional temporal variable (\textit{i.e.}, time step), and $N$ is the length of the trajectory. 

\begin{figure}
    \centering
    \includegraphics[width=0.8\columnwidth]{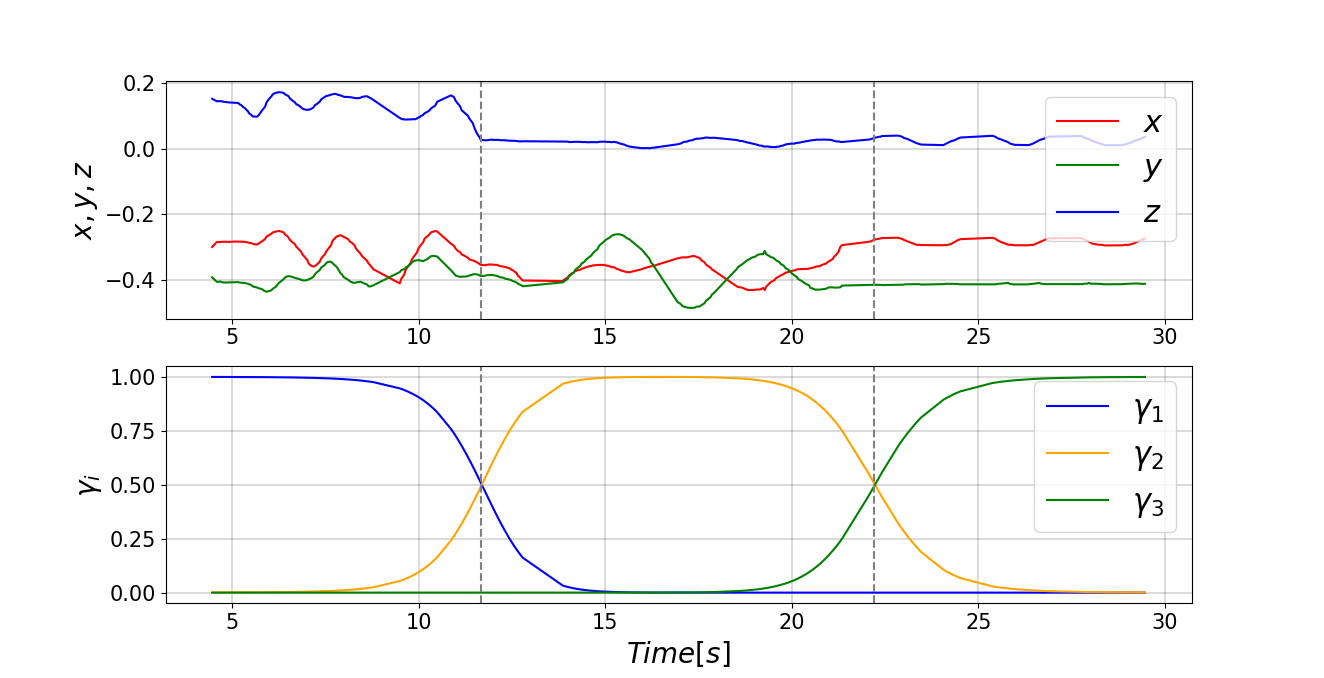}
    \caption{Segmentation framework for Zen Garden Raking. The upper plot represents the mean positional coordinates of the considered keypoints while the lower plot shows segmentation points extracted from weights projected along the time component.}
    \label{fig:projected_weights}
\end{figure}

From the GMM, the segmentation points are detected in temporally overlapping points in-between two consecutive Gaussians; the points are located in all points intersected along the time component of the GMM. However, before detecting these intersections, the mean and the covariance of the time component need to be extracted from the output mean and covariance of the GMM. To this extent, the mean and covariance matrices of the $i^{th}$ Gaussian cluster are represented as 
\begin{equation}
\mu_i = [\mu_{i,t}, \mu_{i,X}], \quad 
\quad
 \Sigma_i  =  \begin{bmatrix} \Sigma_{i,t} & \Sigma_{i,tX} & \\
  \Sigma_{tX,i} & \Sigma_{i,X}\end{bmatrix},
\end{equation}
where $t$ and $X$ refer to the one-dimensional temporal variable and the $D$-dimensional spatial variable in the $(D + 1)$ dimensional trajectory. All intersections between Gaussians are extracted by estimating the weights of Gaussians along the time component. Estimating the weights is defined as
\begin{equation}
    \gamma_i(t) = \frac{w_i\mathcal{N}(t;\mu_{i,t},\Sigma_{i,t})}{\sum_{k=1}^K w_k\mathcal{N}(t;\mu_{k,t},\Sigma_{k,t})}
  ,
\end{equation}
where $i$ and $K$ refer to the indices of Gaussians and the total number of Gaussians, $w_i$ are the output weights of the GMM and $\gamma_i(t)$ represent the weights projected along the time component for each cluster, as seen in Fig. \ref{fig:projected_weights}. Each data point is assigned to a set based on its maximum weight, and the overall trajectory is thus split into $M$ modes, with corresponding data $\mathcal{X}^m$ for the $m^{th}$ mode. 

\subsection{Constraint Fitting}
% Discussion 11.02 Christian & Kevin: may need to try the Hinf fit of constraints, may also be trying the local-feature based constraint description

Given a trajectory segment containing $N$ data points that consists of a single contact mode $m$, the kinematic constraints which are imposed in that mode can be identified. We assume that the kinematic constraints are holonomic constraints in the form $h_m(x, \theta_m) = 0$, where $\theta_m$ are parameters describing the geometry. The constraint $h_m$ is assumed to belong to a set of possible contact primitives \cite{subramani2018}, such as point-on-plane, or line constraint. 

The constraint for a mode is identified by fitting 
\begin{align}
 \min_{\theta_m} & \sum_{x_t \in \mathcal{X}^m} \Vert h_m(x_t, \theta_m)\Vert_2^2 + \beta_m (\theta_m) +Ns & \label{eq:constraint_fit} \\
 \mathrm{s.t.}\,\,\,& \Vert h_m(x_t, \theta_m)\Vert_2^2 \leq s \quad 
\forall x_t \in \mathcal{X}_m&
\end{align}
where $\beta_m$ is a mode dependent regularization term and $s$ a slack variable describing the maximum of $h_m$ which adds an $H_\infty$ norm penalty on $h_m(x_t, \theta_m)$.

\subsubsection{Radius Constraint} The radius constraint describes a point $x\in\mathbb{R}^3$ with a fixed radius $r\in\mathbb{R}$ to a point $x_c\in\mathbb{R}^3$ in base coordinates. All possible values for $x$ are therefore distributed on the surface of a sphere centered at $x_c$.
All points $x_t$ must therefore fulfill the equality constraint $\Vert x - x_c \Vert ^2 = r$. The radius constraint is thus fit by using
\begin{align}
    h_m(x_t, \theta_m) &= r - \Vert x_t - x_c\Vert_2,\quad\quad \beta_m (\theta_m) = k r,
\end{align}
where $k$ is a regularization constant.

\subsubsection{Line on Plane Constraint}\label{sec:lineplaneconst}
The line-on-plane constraint describes an object in line contact with a fixed plane. The object can be translated and rotated along the plane such that the line contact between the object and the plane is never broken. When describing the line with two points $p_0,p_1\in\mathbb{R}^3$ with a fixed relative position to the object and a fixed plane parameterized with its normal vector $n_p\in\mathbb{R}^3$ and offset $d\in\mathbb{R}$ the described constraint can be formulated such that all poses in a data set $\mathcal{X}_m$ that correspond to the described constraint mode must fulfill the equality constraint with
\begin{multline}
h_m(x_t,p_0,p_1,n_p,d) =\\
\begin{bmatrix}
\vert\Vert n_p^T(T(x_t)p_0) \Vert_2 -d \vert \\
\vert\Vert n_p^T(T(x_t)p_1) \Vert_2 - d \vert
\end{bmatrix}
+
\begin{bmatrix}
\Vert n_p^T(p_0-p_1) \Vert_2 \\
\Vert n_p^T(p_0-p_1) \Vert_2
\end{bmatrix},
\end{multline}
\begin{multline}
\beta_m(\theta_m) = \\
k \left(
\left\vert\Vert n_p \Vert_2 \texttt{-}1 \right\vert \texttt{+}
\left\vert\Vert p_0 \texttt{-} p_1 \Vert_2 \texttt{-} k_D \right\vert \texttt{+} 
\Vert p_0 \Vert_2 \texttt{+}
\Vert p_1 \Vert_2
\right),
\end{multline}
where $T(x)\in\mathbb{R}^{4\times 4}$ is the transformation matrix built from the pose $x$, $k$ is a regularization constant and $k_D$ is the desired distance of $p_0$ and $p_1$ in meter.

\subsubsection{Hinge constraint}
A single fixed line in the object coordinate system has a fixed position in world coordinates. The line contact is parameterized with two points $x_{c,0}, x_{c,1}\in\mathbb{R}^3$ in world coordinates and $p_0,p_1\in\mathbb{R}^3$ in object coordinates. Using transformation matrices from the object to the world coordinate system $T(x_t)$ one can express the described relationship as $x_{c,i} = T(x_t) x_{p,i}$, $i=1,2$. The constraint is thus fit with
\begin{align}
h_m(x_t, \theta_m) &= \begin{bmatrix}
    \Vert x_{c,0} - T(x_t) x_{p,0} \Vert_2 \\
    \Vert x_{c,1} - T(x_t) x_{p,1} \Vert_2
\end{bmatrix}, \\
\beta_m (\theta_m) &= 
k (
\vert\Vert p_0 - p_1 \Vert_2 - k_D \vert + 
\Vert p_0 \Vert_2 +
\Vert p_1 \Vert_2
),
\end{align}  
where $k$ and $k_D$ follow the description from \ref{sec:lineplaneconst}. The upper bound $s$ for this constraint by the $H_\infty$ norm is omitted.

\iffalse
\subsubsection{Fixed Point}
A single fixed point in the object coordinate system has a fixed position in world coordinates $x_c\in\mathbb{R}^3$. This constraint point has parameters of $x_p\in\mathbb{R}^3$ in object coordinates and $x_c\in\mathbb{R}^3$ in world coordinates. Using transformations from the object to the world coordinate system $x_t$ one can express the described relationship as $x_c = x_t x_p$. Using multiple objects poses the constraint is thus fit with 
\begin{align}
h_m(x_t, \theta_m) &= \Vert x_c - x_t x_p \Vert_2,\quad \beta (\theta_m) = 0.
\end{align}
The upper bound $s$ for this constraint by the $H_\infty$ norm is again omitted in the optimization process.
\fi

\section{Online Control}
This section describes how the contact models can be employed in the online control of the robot. 

\subsection{Contact Detection \label{sec:contact_detection}}
Given a constraint $h_m(x,\theta_m)\in\mathbb{R}^{n_m}$, the forces which would enforce that constraint can be found as 
\begin{equation}
 F = J_m^T(x)\lambda + \epsilon, \label{eq:forces}
\end{equation}
where force $F\in\mathbb{R}^n$, Jacobian $J_m(x) = \frac{\partial h_m(x, \theta_m)}{\partial x}\in\mathbb{R}^{n_m \times n}$,  $\lambda\in\mathbb{R}^{n_m}$ are Lagrange multipliers which enforce the constraint and are not measured, and $\epsilon$ additional forces. The additional forces $\epsilon$ can be from gravitational or inertial forces of the payload, friction in contact, or other errors. The forces $F$ are expressed in the coordinate system of the pose $x$. 

In order to efficiently detect the active constraint in the online framework we need a robust metric that could scale properly for constraints with higher dimensions ($n_m>1$). We formulate the problem as a least-squares problem to exploit the closed-form solution through the pseudoinverse $J_m(x)^\dag$ of the constraint Jacobian as
\begin{equation}
\epsilon(F, x) = \Vert F - J_m(x)^\dag J_m(x) F\Vert_2. 
\end{equation}
This residual tells us the amount of measured force that is not explained by the constraint and gives a metric for online detection: the constraint with a lower residual is selected as the current constraint. 

As the measured wrench vector $F_t$ is measured at the TCP, it needs to be transformed to match the orientation of the coordinate system of the pose $x$, expressed with $XYZ$ extrinsic Euler angles. We transform the wrench into a frame at the origin of the object frame and with the same orientation as the base frame. The standard transform of the full wrench vector with respect to a new reference frame can be found in \cite{bo2001}. 

To reduce false contact detection from the acceleration of the payload we also model the free space as a type of contact mode, but we do not detect free space based on the force residual. Free space detection is done when $ \Vert F_\tau \Vert < \overline{F} \quad \forall\tau\in [t-w, t] $, where $\overline{F}\in \mathbb{R}$ is a threshold, $w$ a properly tuned time window and $\vert F_\tau \vert$ is the norm of the measured wrench at TCP for each time step.

% 09.02.23: Discussion Filippo & Kevin: Not clear how to scale to n_m>1. 
% Detection of contact is done by comparing the measured force $F_t$ with the force direction indicated by constraints $J_m(x)$. In the simplest case, $J_m\in\mathbb{R}^n$ (\textit{i.e.}, point contacts) and a simple similarity metric is \begin{equation} s_m(F_t,x_t) = F_t^TJ_m(x_t,\theta_m) \end{equation} which indicates the component of measured forces in the direction of the constraint normal.

\subsection{Constraint-aware Control}
The constraint information can also be directly used in the control strategy. We use a lower-level admittance controller, such that
\begin{eqnarray}
M\ddot{x}^r_t + D\dot{x}^r_t + K\left(x^d_t-x^r_t\right) = F^d_t-F_t,  \label{eq:admittance}
\end{eqnarray}
where $F^d\in\mathbb{R}^6$ is the force reference, $F_t$ the measured force in robot TCP, $x^d_t$ the desired pose of the robot, $x^r_t$ the robot pose and $M$, $D$, $K\in\mathbb{R}^{6\times 6}$ the mass, damping, and stiffness admittance parameters. All quantities in \eqref{eq:admittance} are expressed in the TCP frame, and the matrices $M$, $D$, and $K$ are diagonal.  The following sections propose how the reference signals $F^d_t$ and $x^d_t$ can be adjusted online according to the mode information.

\subsubsection{Contact-triggered Control}
At each time step $t$, we infer the current mode $m_t$ as shown in Section \ref{sec:contact_detection}. We then suppose that each mode has its own motion strategy as
\begin{equation}
    x^d_t = \pi_{m_t}(x^r_t), 
\end{equation}
where $\pi_{m_t}$ is a mode-specific controller which produces a desired pose $x^d_t$. 

\subsubsection{Making and Maintaining Contact}
Many constraints are unilateral constraints more accurately considered as $h(x)\geq 0$, for example sliding on the contact of a surface. This means the constraint can be lost by moving away from the constraint, which may be undesired for the task and limits the contact detection. 

Additionally, the pose where contact is entered may vary. Additional motion in the direction of the next constraint, as given by the Jacobian of the next constraint, can help ensure that contact is made. This is done by adding to the desired contact wrench.

We add a small desired force to the robot force controller as
\begin{equation}
    F^d_t = T_w\left(J^T_{m_t}(x_t)\alpha + J^T_{m_{t+1}}(x_t)\alpha_+\right),
\end{equation}
where $T_w$ transforms the wrench from object coordinates to TCP, $m_t$ is the current mode,  $m_{t+1}$ the next contact, $\alpha\in\mathbb{R}^{n_{m_t}},\, \alpha>0$ specifies the force to apply to maintain the current constraint and $\alpha^{+}$ the force to approach the next constraint.

\section{Validation}
The approach is validated on a UR16e robot, which has an integrated F/T sensor at the flange and a Realsense D435 camera. The CasADi framework is used for parameter fitting and calculating Jacobians \cite{andersson2019}. The admittance control is implemented via ROS \cite{scherzinger2019}. The code is available at \url{https://github.com/khaninger/contact_monitoring}.  

\subsection{Cable pulling}
\begin{figure}
    \centering
    \includegraphics[width=0.65\columnwidth]{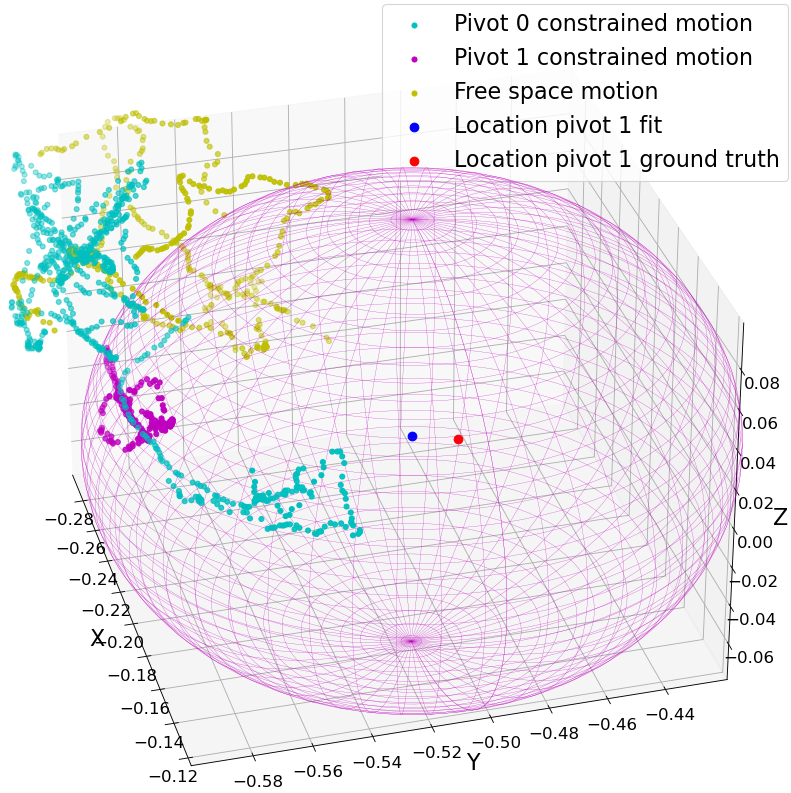}
    \caption{Segmentation of cable pulling data with fit parameters for front pivot}
    \label{fig:plot_cable}
\end{figure}

The data collected for the radius constraint parameter estimation consists of 3D positions of a plug that is constrained with a cable that is fixed via two different pivots to a workbench. The demonstration data is shown in Fig. \ref{fig:plot_cable}, where the observed plug positions are segmented into: free space (yellow), constraint by cable fixed at pivot 0 (cyan) and pivot 1 (magenta). The segmentation process has some noise in detecting the transitions but provides sufficient accuracy for the constraint fitting.  
 
The location of the fit pivot 1 and its ground truth is shown in blue respectively red. Additionally, the sphere for the estimated radius around pivot 1 is shown. The fit constraint results in an error of $3.2$cm with respect to the measured pivot, where the position is measured at height of the workbench with the camera from pixel space just as the keypoints are measured.

Using the fit constraints, the controller is applied as seen in the attached video. The parameters used for constraint detection are $\overline{F}=6$, window $w=8$, and constraint forces are applied of $\alpha = 12$, $\alpha^{+}=6$.  The stiffness parameters of the impedance controller are $K=[250, 250, 550, 20, 20, 20]$. The motion strategy $\pi_{m_t}(x_t)$ finds the closest point in the demonstration trajectory, generates $x^d_t$ by transforming from the object frame to TCP, and advances along the trajectory when the distance to the current $x^d_t$ is less than $1$ cm. The demonstration trajectory moves from free space, pulling tight with the pivot 0 constraint, then moves directly to enter the pivot 1 constraint, similar to that seen in Fig. \ref{fig:plot_cable}.

\begin{figure}
    \centering
    \subfloat[Normal cabling process]{\includegraphics[width=0.75\columnwidth]{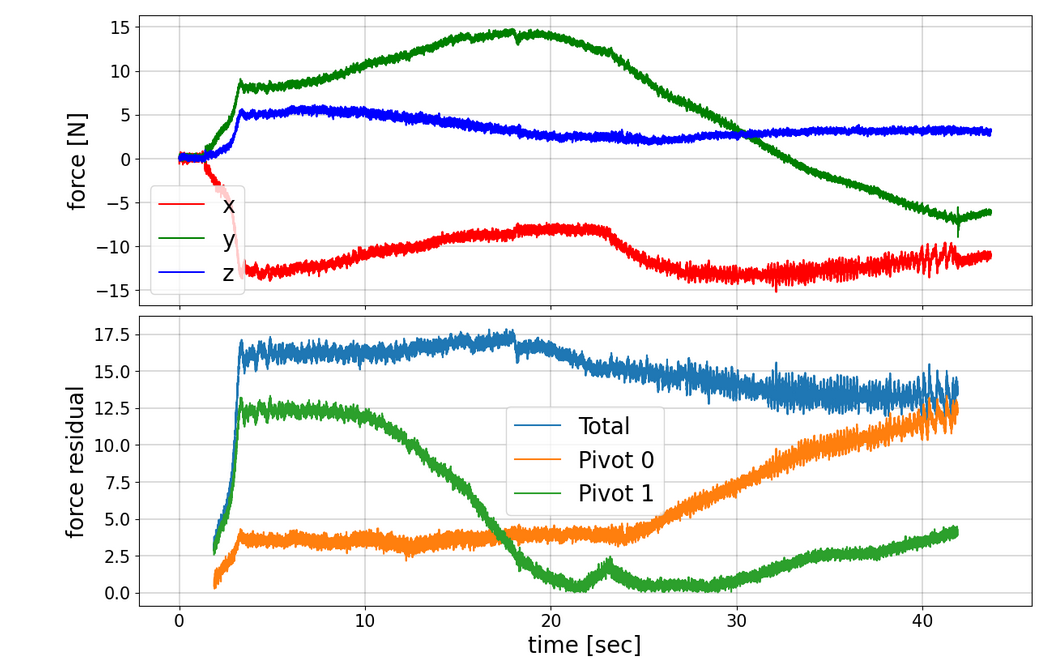}}\\
    \subfloat[Disturbed cabling process]{\includegraphics[width=0.75\columnwidth]{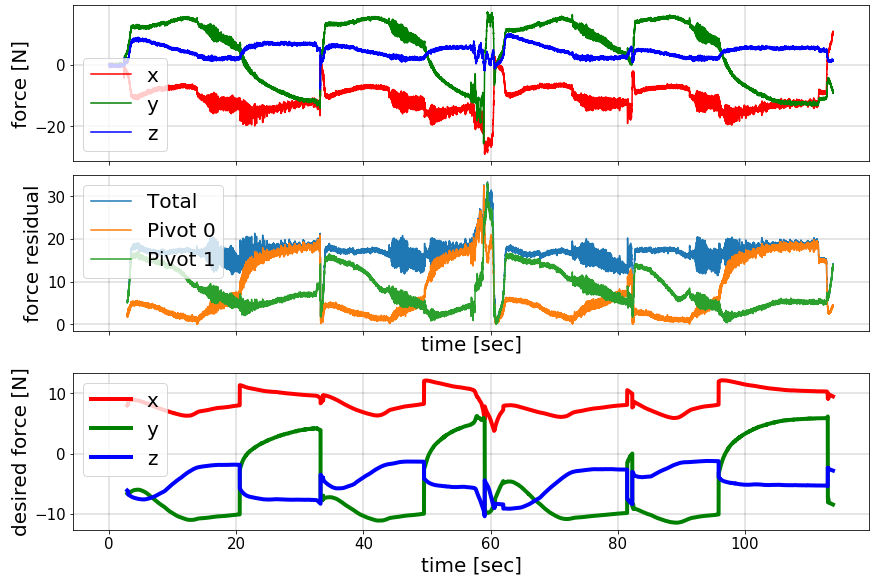}}
    \caption{Forces and contact residual in the cabling task}
    \label{fig:online_cable}
\end{figure}

Results of the online detection can be seen in the attached video and Fig. \ref{fig:online_cable}(a), where the residual per constraint and forces are shown. Near the starting pose, the residual between pivot 0 and pivot 1 can be clearly distinguished - at that point these constraints are in different directions. However, as the robot moves near when the two constraints are in a similar direction (around $20$ sec), the residuals become more similar, and a few centimeters of error in the mode detection were noticed.  This is attributed to the constraints being in a similar direction - the vectors $J_0$ and $J_1$ for pivot 0 and 1 are almost identical, and distinguishing the forces is difficult.  Similar difficulty in distinguishing certain constraints has been noted by other authors \cite{subramani2018, debus2005}.

To validate the robustness of detecting changes in contact, the cable is unhooked by hand from pivot 1 multiple times. Additionally, at 60 seconds the robot is manually perturbed 20 cm from the path and at 80 seconds the cable slips on its own. The robot is able to recover from all these perturbations, as seen in the attached video. The force residuals can be seen in Fig. \ref{fig:online_cable}(b), where it can be seen that this online change is robustly detected, and the robot can recover with the mode dependent control. Furthermore, adding the desired force along the constraint direction allows the robot to keep the cable tight while executing the task, which does not always occur when using admittance control with $F^d_t=0$.

\subsection{Zen Garden Raking}
This task involves a rake over a planar surface, with the additional constraint of reaching the edge of the raking area.  These two constraints are extracted from demonstrations, where the pose of the rake is tracked as the rake moves in free space, line contact with the workbench, then a hinge constraint with the border of the raking area is met. The rake pose is tracked with four keypoints, and a keypoint is saved only when all four are observed, and the residual error of the correspondence to the reference keypoints is less than $3$ mm. 

\begin{figure}
    \centering
    \includegraphics[width=0.75\columnwidth]{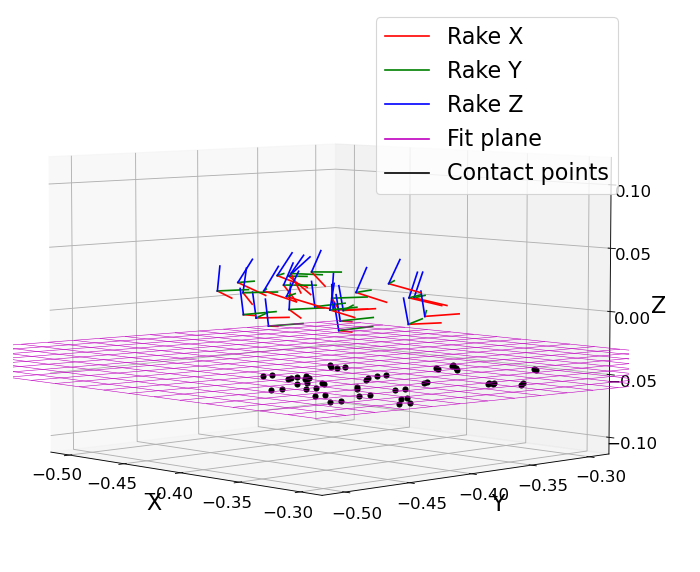}
    \caption{Poses from rake data with fit parameters for the plane contact, with corresponding contact points $p_0,\, p_1$}
    \label{fig:plot_rake}
\end{figure}

A subset of the observed poses of the rake in contact including the fit plane and projected contact points for all shown poses are visualized in Fig. \ref{fig:plot_rake}, along with the points of the identified contact point on the line constraint (black points) and the fit plane location (purple). 

\begin{figure}
    \centering
    \includegraphics[width=0.75\columnwidth]{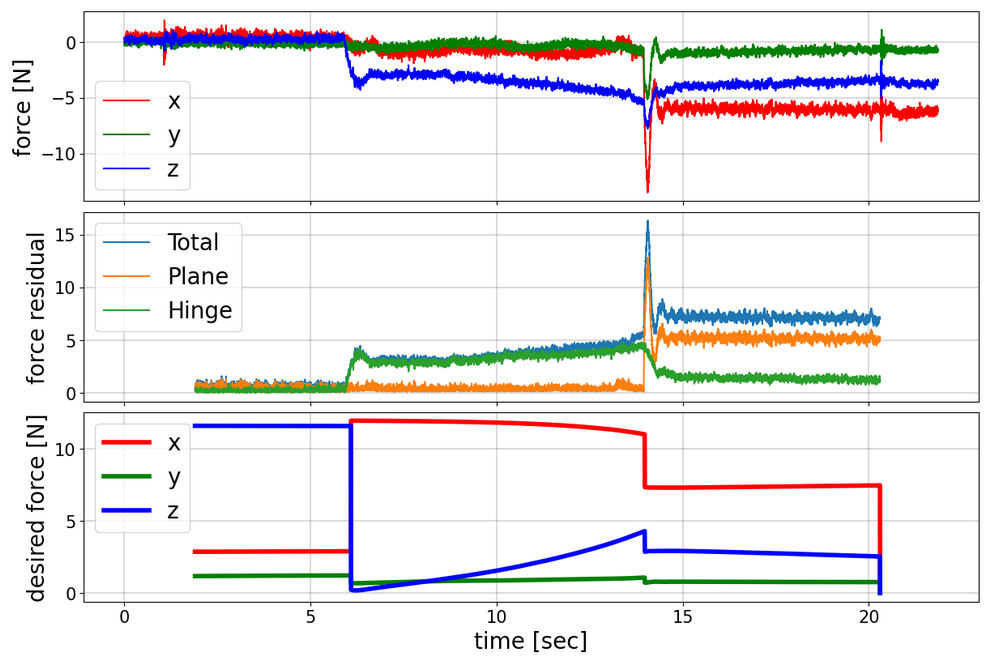}
    \caption{Measured forces in TCP, constraint residual, and desired force $F^d_t$ in the rake task}
    \label{fig:online_rake}
\end{figure}

The results of the live control experiments on the rake can be seen in Fig. \ref{fig:online_rake} and the attached video. The parameters used are $\overline{F}=1.4$, $w=6$, $\alpha = 4$ and $\alpha^+ = 6$. The first contact, just after 5 seconds, results in an overshoot, but the residual for the plane  constraint approaches zero with no overshoot, indicating most of the collision forces are aligned with the expected constraint. The contact with the second constraint (hinge) results in an impulse in the forces, but again this impulse does not appear in the residual for the hinge constraint, indicating the impulse is occurring primarily in the subspace expected by the constraint.

\section{Conclusion}
The paper has proposed a pipeline for learning constraints from the passive observation of markerless human demonstrations. A general constraint fitting process is introduced, and several example constraints are validated from real data. The accuracy of a keypoint-based pose estimation, GMM-based clustering, and constraint model identification is shown to have sufficient accuracy for some contact-rich tasks. These constraint models are shown to result in effective online detection of constraint from F/T information, provided the Jacobian between them is sufficiently distinct. The usefulness of this information was demonstrated by simple mode-dependent trajectory following with a desired force which maintains the current constraint.  

\bibliographystyle{IEEEtran}
\bibliography{biblio}

\end{document}